\documentclass[10pt,twocolumn,letterpaper]{article}

\usepackage{cvpr}              

\usepackage[dvipsnames]{xcolor}

\usepackage{graphicx}
\usepackage{amsmath}
\usepackage{amssymb}
\usepackage{booktabs,eucal}

\usepackage{ upgreek } 

\usepackage{graphicx}
\usepackage{subcaption}

\usepackage{multirow} 

\definecolor{cvprblue}{rgb}{0.21,0.49,0.74}
\usepackage[pagebackref,breaklinks,colorlinks,citecolor=cvprblue]{hyperref}

\usepackage[capitalize]{cleveref}
\crefname{section}{Sec.}{Secs.}
\Crefname{section}{Section}{Sections}
\Crefname{table}{Table}{Tables}
\crefname{table}{Table}{Tables}

\usepackage[symbol]{footmisc}

\def\paperID{****} 
\def\confName{CVPR}
\def\confYear{2024}

\title{FaceStudio: Put Your Face Everywhere in Seconds}

\author{Yuxuan Yan$^{\ast}$ 
\quad 
Chi Zhang$^{\ast\dagger}$ 
\quad Rui Wang \quad Yichao Zhou \quad Gege Zhang \quad \\ Pei Cheng  \quad Bin Fu   \quad Gang Yu \vspace{0.3em} \\
{ Tencent}  \\
{\normalsize \{yuxuanyan,~johnczhang,~raywwang, ~yichaozhou, ~gretazhang, ~peicheng,~brianfu, ~skicyyu\}@tencent.com} \\
\url{https://icoz69.github.io/facestudio/}
}

\begin{document}

\makeatletter
\let\@oldmaketitle\@maketitle%
\renewcommand{\@maketitle}{\@oldmaketitle%
 \centering
    \includegraphics[width=1\textwidth]{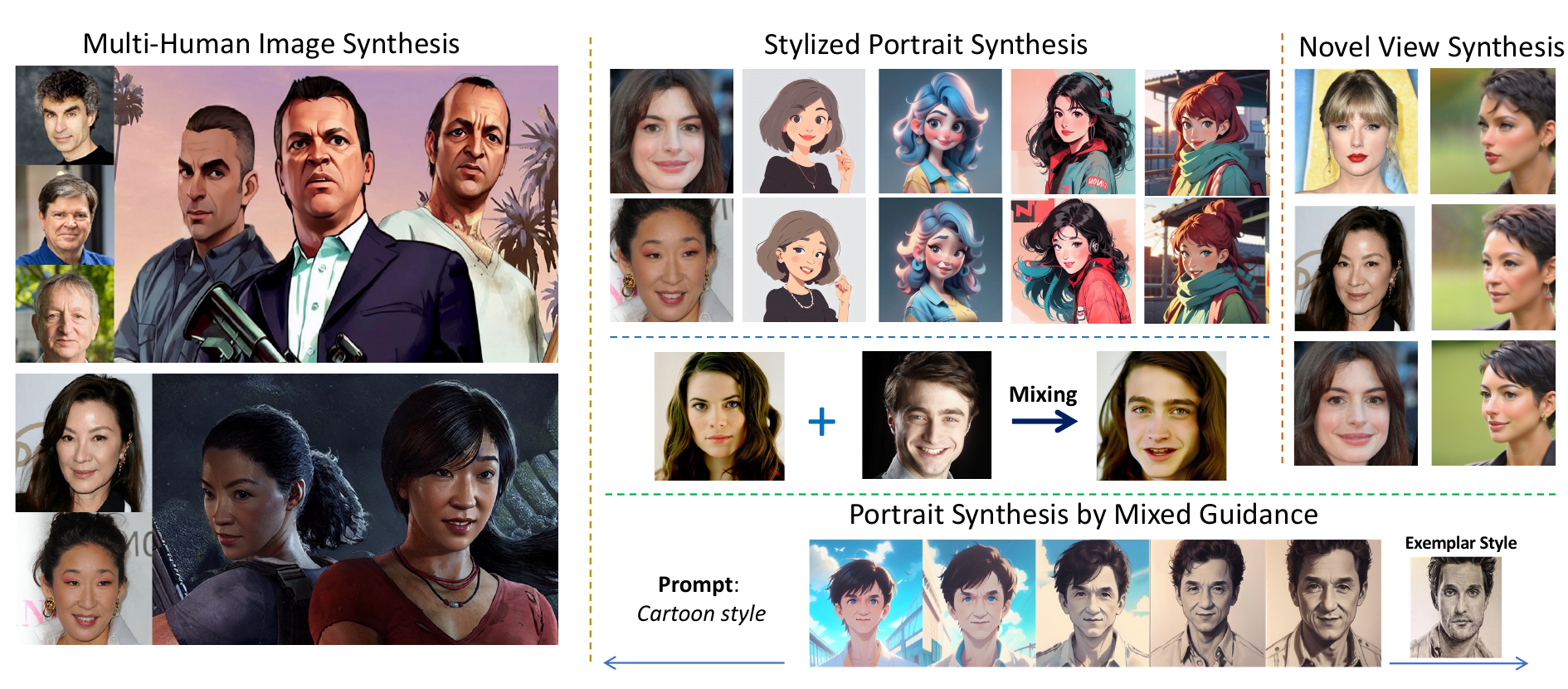}
    \vskip -0.5em
     \captionof{figure}{ 
     \textbf{Applications of our proposed framework for identity-preserving image synthesis.} Our method can preserve the subject's identity in the synthesized images with high fidelity.
     }
    \label{fig:teaser}
    \bigskip}           
\makeatother

\maketitle

\footnotetext[1]{Equal contributions.}
\footnotetext[2]{Corresponding Author.}


\begin{abstract}
This study investigates identity-preserving image synthesis, an intriguing task in image generation that seeks to maintain a subject's identity while adding a personalized, stylistic touch. Traditional methods, such as Textual Inversion and DreamBooth, have made strides in custom image creation, but they come with significant drawbacks. These include the need for extensive resources and time for fine-tuning, as well as the requirement for multiple reference images. To overcome these challenges, our research introduces a novel approach to identity-preserving synthesis, with a particular focus on human images. Our model leverages a direct feed-forward mechanism, circumventing the need for intensive fine-tuning, thereby facilitating quick and efficient image generation. Central to our innovation is a hybrid guidance framework, which combines stylized images, facial images, and textual prompts to guide the image generation process. This unique combination enables our model to produce a variety of applications, such as artistic portraits and identity-blended images. Our experimental results, including both qualitative and quantitative evaluations, demonstrate the superiority of our method over existing baseline models and previous works, particularly in its remarkable efficiency and ability to preserve the subject's identity with high fidelity.

\end{abstract}

\section{Introduction}

In recent years, artificial intelligence (AI) has driven significant progress in the domain of creativity, leading to transformative changes across various applications. Particularly, text-to-image diffusion models~\cite{stablediffusion,dalle2} have emerged as a notable development, capable of converting textual descriptions into visually appealing, multi-styled images. Such advancements have paved the way for numerous applications that were once considered to be beyond the realms of possibility.

However, despite these advancements, several challenges remain. One of the most prominent is the difficulty faced by existing text-to-image diffusion models in accurately capturing and describing an existing subject based solely on textual descriptions. This limitation becomes even more evident when detailed nuances, like human facial features, are the subject of generation. Consequently, there is a rising interest in the exploration of identity-preserving image synthesis, which encompasses more than just textual cues. In comparison to standard text-to-image generation, it integrates reference images in the generative process, thereby enhancing the capability of models to produce images tailored to individual preferences.

In pursuit of this idea, various methods have been proposed, with techniques such as  DreamBooth~\cite{ruiz2023dreambooth} and Textual inversion~\cite{gal2022image} leading the way. They primarily focus on adjusting pre-trained text-to-image models to align more closely with user-defined concepts using reference images. However, these methods come with their set of limitations. The fine-tuning process, essential to these methods, is resource-intensive and time-consuming, often demanding significant computational power and human intervention. Moreover, the requirement for multiple reference images for accurate model fitting poses additional challenges.

In light of these constraints, our research introduces a novel approach focusing on identity-preserving image synthesis, especially for human images. Our model, in contrast to existing ones, adopts a direct feed-forward approach, eliminating the cumbersome fine-tuning steps and offering rapid and efficient image generation. Central to our model is a hybrid guidance module, which guides the image generation of the latent diffusion model. This module not only considers textual prompts as conditions for image synthesis but also integrates additional information from the style image and the identity image. By employing this hybrid-guided strategy, our framework places additional emphasis on the identity details from a given human image during generations. 
To effectively manage images with multiple identities, we develop a multi-identity cross-attention mechanism, which enables the model to aptly associate guidance particulars from various identities with specific human regions within an image.

Our training method is intuitive yet effective. Our model can be easily trained with human image datasets. By employing images with the facial features masked as the style image input and the extracted face as the identity input, our model learns to reconstruct the human images while highlighting identity features in the guidance.
After training, our model showcases an impressive ability to synthesize human images that retain the subject's identity with exceptional fidelity, obviating the need for any further adjustments.
 A unique aspect of our method is its ability to superimpose a user's facial features onto any stylistic image, such as a cartoon, enabling users to visualize themselves in diverse styles without compromising their identity. Additionally, our model excels in generating images that amalgamate multiple identities when supplied with the respective reference photos. Fig.~\ref{fig:teaser} shows some applications of our model.

This paper's contributions can be briefly summarized as follows:
 \begin{itemize}
\item  We present a tuning-free hybrid-guidance image generation framework capable of preserving human identities under various image styles.

\item We develop a multi-identity cross-attention mechanism, which exhibits a distinct ability to map guidance details from multiple identities to specific human segments in an image.

\item   We provide comprehensive experimental results, both qualitative and quantitative, to highlight the superiority of our method over baseline models and existing works, especially in its unmatched efficiency.

\end{itemize}

\begin{figure*}[t]
\centering
\includegraphics[width=0.95\textwidth]{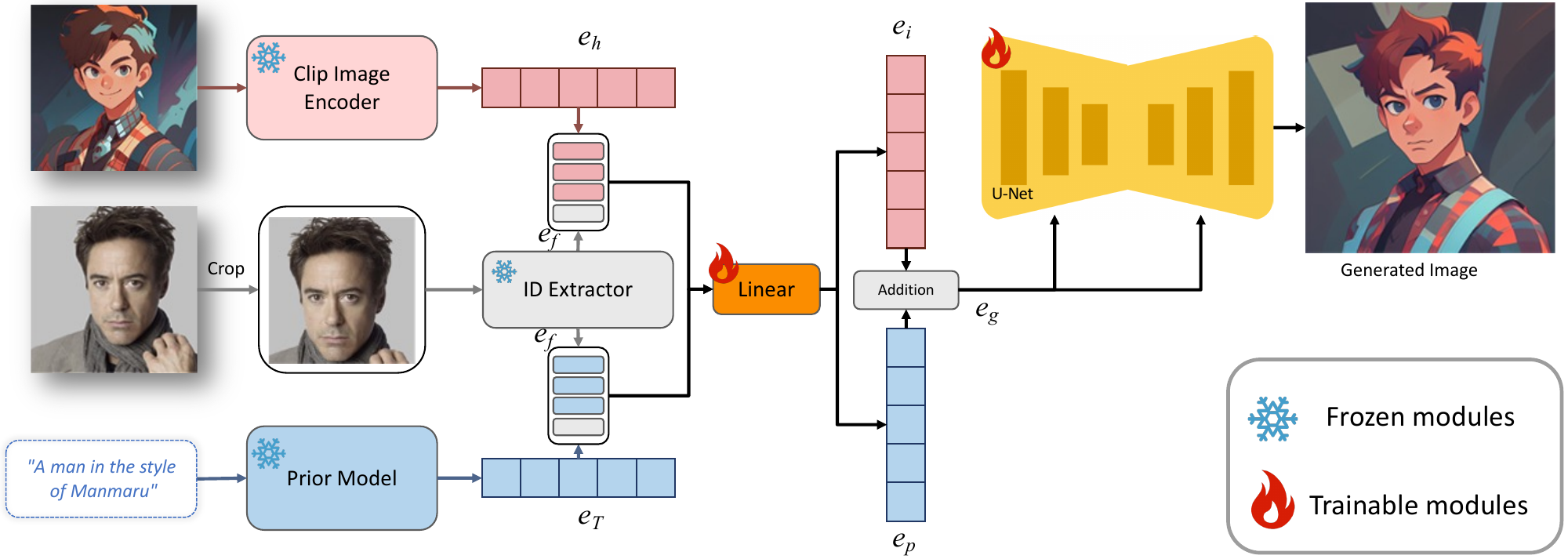}
\vspace{-0.5 em}
\caption{\textbf{Hybrid-Guidance Identity-Preserving Image Synthesis Framework.} Our model, built upon StableDiffusion, utilizes text prompts and reference human images to guide image synthesis while preserving human identity through an identity input. } 
\label{fig:main}
\vspace{-1.5em}
\end{figure*}

\section{Related Work}
\textbf{Text-to-image diffusion models.}
Diffusion models have recently come to the forefront of generative model research. Their exceptional capability to generate high-quality and diverse images has placed them at a distinct advantage over their predecessors, namely the GAN-based and auto-regressive image generation models. 
This new generation of diffusion models, having the capacity to produce state-of-the-art synthesis results, owes much of its success to being trained on image-text pairs data at a billion scale.
The integration of textual prompts into the diffusion model serves as a vital ingredient in the development of various text-to-image diffusion models. Distinguished models in this domain include GLIDE~\cite{nichol2021glide}, DALL·E 2~\cite{dalle2}, Imagen~\cite{imagen}, and StableDiffusion~\cite{stablediffusion}. These models leverage text as guidance during the image generation process. Consequently, they have shown considerable proficiency in synthesizing high-quality images that closely align with the provided textual description.
Compared to previous efforts~\cite{perov2020deepfacelab, li2019faceshifter, wang2021hififace, zhu2021one} on face image synthesis based on Generative Adversarial Networks (GANs)~\cite{karras2018stylebased}, diffusion models exhibit greater stability during training and demonstrate enhanced capability in effectively integrating diverse forms of guidance, such as texts and stylized images.

However, textual descriptions for guidance, while immensely useful, often fall short when it comes to the generation of  complex and nuanced details that are frequently associated with certain subjects. For instance, generating images of human faces proves challenging with this approach. While generating images of celebrities might be feasible due to the substantial presence of their photographs in the training data that can be linked to their names, it becomes an uphill task when it comes to generating images of ordinary people using these text-to-diffusion models.

\textbf{Subject-driven image generation.}
Subject-driven image generation seeks to overcome the limitations posed by text-to-image synthesis models.
Central to this novel research area is the inclusion of subject-specific reference images, which supplement the textual description to yield more precise and personalized image synthesis.
To this end, several optimization-based methods have been employed, with popular ones being Low-Rank Adaptation (LoRA), DreamBooth~\cite{ruiz2023dreambooth}, Textual Inversion~\cite{gal2022image}, and Hypernetwork~\cite{hypernetwork}. These methods typically involve fine-tuning a pre-trained text-to-image framework or textual embeddings to align the existing model with user-defined concepts, as indicated by a set of example images. There are several other methods derived from these works such as Unitune~\cite{valevski2022unitune}, HyperDreamBooth~\cite{ruiz2023hyperdreambooth}, EasyPhoto~\cite{wu2023easyphoto}, FaceChain~\cite{liu2023facechain}, \textit{etc}. ~\cite{jeong2023zero, yuan2023inserting, alaluf2023neural}.
However, these methods pose some challenges. For one, they are time-consuming due to the need for model fine-tuning, which also requires multiple reference images to achieve accurate fitting. Overfitting also becomes a potential issue with these optimization-based methods, given their heavy reliance on example images.
In response to these challenges, recent studies have proposed various improved methods. DreamArtist~\cite{dong2022dreamartist}, for instance, mitigates the problem of overfitting by incorporating both positive and negative embeddings and jointly training them. Similarly, E4T~\cite{gal2023encoder} introduced an encoder-based domain-tuning method to accelerate the personalization of text-to-image models, offering a faster solution to model fitting for new subjects.
There are large numbers of similar methods which are encoder-based including Composer~\cite{huang2023composer},
ProFusion~\cite{zhou2023enhancing}
MagiCapture~\cite{hyung2023magicapture},
IP-Adapter~\cite{ye2023ip},
ELITE~\cite{wei2023elite},
DisenBooth~\cite{chen2023disenbooth},
Face0~\cite{valevski2023face0},
PhotoVerse~\cite{chen2023photoverse},
AnyDoor~\cite{chen2023anydoor},
SingleInsert~\cite{wu2023singleinsert}, \textit{etc}. ~\cite{jia2023taming, ma2023subject, xu2023prompt}.

Alongside optimization-based methods, a number of tuning-free methods have been concurrently proposed, such as InstantBooth~\cite{shi2023instantbooth}. This method converts input images into a textual token for general concept learning and introduces adapter layers to obtain rich image representation for generating fine-grained identity details. However, it comes with the drawback of having to train the model separately for each category.
Bansal et al.\cite{bansal2023universal} put forward a universal guidance algorithm that enables diffusion models to be controlled by arbitrary guidance modalities without the need to retrain the network. Similarly, Yang et al.\cite{yang2023paint} propose an inpainting-like diffusion model that learns to inpaint a masked image under the guidance of example images of a given subject. Similar methods leveraged by inversion-based personalization including Cao et al.~\cite{cao2023masactrl},
Han et al.~\cite{han2023improving},
Gu et al.~\cite{gu2023photoswap},
Mokady et al.~\cite{mokady2023null}.
Besides, Controlnet~\cite{zhang2023adding} is also an effective way to personalize. In addition, there are some solutions that use editing to achieve the purpose of maintaining identity, such as SDEdit~\cite{meng2021sdedit},
Imagic~\cite{kawar2023imagic}, \textit{etc}. ~\cite{hertz2022prompt, tewel2023key}.
Based on all the solutions mentioned above, the method of combining multiple objects while maintaining identity is also widely used in Break-a-scene~\cite{avrahami2023break},
FastComposer~\cite{xiao2023fastcomposer},
Cones~\cite{liu2023cones} and
MultiConceptCustomDiffusion~\cite{kumari2023multi}.

Within the text-to-image community, one specific area of interest that has garnered significant attention is the generation of human images. Human-centric image generation, owing to its vast applicability and widespread popularity, constitutes a substantial proportion of posts in the community, such as Civitai~\cite{Civitai}. 
In light of these considerations, our work primarily focuses on the preservation of human identity during the image generation process.

\textbf{Few-shot Learning.} It is a common practice in the field of few-shot learning literature to convert optimization-based tasks into a feed-forward approach, which can significantly reduce computation time. Numerous previous studies in the realm of few-shot learning have adopted this approach to address computer vision tasks, including but not limited to classification~\cite{zhang2020deepemd,zhang2022deepemd,zhang2021few,zhang2021meta}, detection~\cite{yang2022efficient}, and segmentation~\cite{zhang2019pyramid,zhang2019canet,liu2020crnet,liu2022crcnet,chen2020compositional,liu2021few,lai2022tackling}. In this research, our objective aligns with this approach as we aim to integrate identity information into the generative process through a feed-forward module.
\begin{figure}
   \centering
   \begin{subfigure}{0.45\textwidth}
       \centering
       \includegraphics[width=0.9\linewidth]{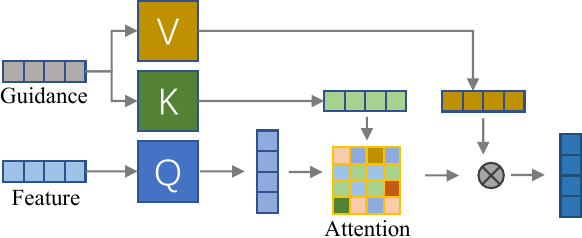}
       \caption{Baseline cross-attentions within the StableDiffusion model.}
       \label{fig:sub1}
   \end{subfigure}
   \hfill
   \begin{subfigure}{0.45\textwidth}
       \centering
       \includegraphics[width=0.9\linewidth]{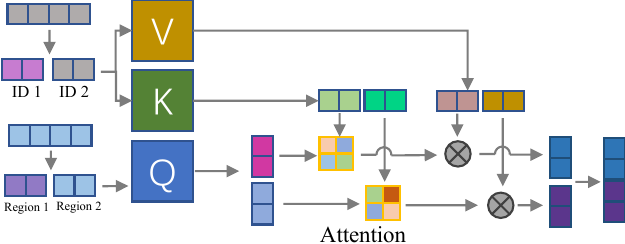}
       \caption{Enhanced cross-attentions optimized for multi-identity synthesis.}
       \label{fig:sub2}
   \end{subfigure}
   \caption{  Comparison between standard cross-attentions in single-identity modeling \textbf{(a)} and the advanced cross-attentions tailored for multi-identity integration \textbf{(b)}. }
   \label{fig:cross_attention}
   \vspace{-1 em}
\end{figure}

\section{Method}
In this section, we present the design and functionalities of our novel framework. Our method fundamentally builds on StableDiffusion~\cite{stablediffusion},  with several pivotal modifications, especially in the condition modules catering to hybrid-guidance image generation. 
We start by elaborating on our hybrid guidance design in the proposed condition module. Following that, we delve into the mechanism for managing multiple identities within images. Lastly, we discuss the training strategy of our models.
The overview of our model structure is shown in Fig.~\ref{fig:main}.

\subsection{Hybrid Guidance Strategy}

\textbf{Disentangling Identities from Style Images}
Given our research's main aim is to keep human identities intact during image generation under various styles, it is indispensable to extract salient identity features from human images for conditional image generation. 
Building on this understanding, our first step is setting up an image-condition guidance module, which aims to take stylized human pictures as the main data for the condition modules. 
We additionally incorporate a human identity input in this module, which works with face pictures to guide the synthesis of human images. Intuitively, the stylized human pictures also referred to as \textit{style images}, specify image content and style, while the identity input provides fine-grained identity guidance.
More specifically,  we deploy the CLIP vision encoder~\cite{clip} to process the human images \(I_h\), resulting in \(e_h=\mathtt{CLIP_V} (I_h)\). Concurrently, the Arcface model~\cite{deng2018arcface} attends to a face image \(I_f\), leading to \(e_f=\mathtt{Arcface} (I_f)\). These two derived embeddings are then combined, and a linear layer processes the combined data: \(e_i=\mathtt{Linear} (e_h || e_f)\), where \(||\) denotes the concatenation function. This design effectively extracts and disentangles the identity-related representations from the overall human image content. By doing so, it equips the model with the capability to specifically focus on human identities during the synthesis process. An intrinsic advantage of this separated design is the flexibility it offers. After training, by simply swapping the facial image in the condition module, we can seamlessly modify the identity within a human image, ensuring adaptability to diverse requirements.

\textbf{Incorporating Textual Guidance}
To further allow textual prompts as guidance for conditional image generation,  a  prior model~\cite{dalle2} is employed to translate textual descriptions \(T\) into image embeddings \(e_T\). To achieve this, the prior model is pre-trained to map the CLIP text embeddings to their corresponding CLIP vision embeddings, resulting in \(e_T=\mathtt{Prior} (\mathtt{CLIP_T} (T))\). Following this, a linear layer is deployed to integrate the textual guide with the identity information, formulated as \(e_p=\mathtt{Linear} (e_T || e_f)\).  
Given that the prior model's output shares the embedding space with the CLIP vision encoder's output, \textit{both branches in our framework use shared linear layer parameters for identity fusion.}
  With these steps, the model is equipped with dual guiding mechanisms for image generation: human photographs under various styles and text prompts, both enriched with identity information. The two types of guidance are then merged linearly, using a hyper-parameter \(\alpha\), which  provides the final guidance embedding \(e_g\): \(e_g= {\alpha}e_i + (1-\alpha)e_p \).
  Finally, this guidance embedding is fused into the U-Net with cross-attentions, as is done in StableDiffusion~\cite{stablediffusion}.

\subsection{Handling Multiple Identities}

Our design can adeptly fuse multiple identities when given their respective face photos. This merging is done by blending their face embeddings, presented as \(e_f=\sum_{i=1}{\beta^i e^i_f}\), where \(e^i_f\) is the facial embedding of the $i$th identity, and \(\beta^i\) is a weight factor.
Yet, a challenge arises when we intend to superimpose varying identities onto multiple humans within a single image. Given the standard process in StableDiffusion~\cite{stablediffusion}, guidance information, \(e_g\), is fused into  the intermediate U-Net feature maps, \(F\), using a cross-attention layer, represented as  
\begin{equation}
\hat{F}=\mathtt{Attention}(Q, K, V).
\end{equation}
Here, \(Q\) originates from flattened intermediate features \(F\), while \(K\) and \(V\) come from the guidance embedding \(e_g\).
In situations with several humans in the content images, we aim for each human's generation to reference unique identity data. If there are \(N\) humans and \(N\) identities, with the aim of a 1-to-1 identity match, our approach ensures that features in \(F\) from the \(i\)th human region solely access information from the \(i\)th identity. This is denoted as 
\begin{equation}
\hat{F^i}=\mathtt{Attention}(Q^i, K^i, V^i), 
\end{equation}
where \(Q^i\) is obtained from the specific features of the \(i\)th human, and \(K^i\) and \(V^i\) are derived from the guidance vector of the \(i\)th identity. 
This feature selection operation can be easily implemented with the help of an off-the-shelf human instance segmentation model. An illustration of our design is shown in Fig.~\ref{fig:cross_attention}.
The strength of our design lies in its precision in managing multiple identities, ensuring that each identity aligns with its corresponding human figure in the content image. 
For features in non-human regions, they can randomly choose guidance embeddings for guidance, since all guidance embeddings contain information about the stylized human image, with differences only in the identity inputs.
Alternatively, the latent embeddings in these regions can be fixed, similar to the approach used in StableDiffusion~\cite{stablediffusion} for inpainting tasks.

\subsection{Training}
We train our model with a human image reconstruction task, conditioned on the input human image and the identity input.
Specifically, raw images with the face region masked serve as the stylized human image, while the cropped face from the same image acts as the identity input. This strategy effectively separates identity information from the overall image content for guided image generation.
Our model is optimized by only using the image-conditioned branch in our condition module, as the linear layer in the text-conditioned branch shares the parameters with the one in the image-condition branch.
This obviates the need for text annotations for human images, which are often hard to acquire on a large scale. We keep the parameters of both the CLIP vision encoder and the Arcface model frozen, focusing our optimization on the newly introduced linear layer and the U-Net model.
In line with StableDiffusion~\cite{stablediffusion}, our U-Net model, denoted as $\varepsilon_\theta()$, is trained to denoise latent representations produced by a pre-trained VAE encoder $\mathcal{E}()$. This is captured by:
\begin{equation}
L_{DM} := \mathbb{E}_{\mathcal{E}(x), \epsilon \sim \mathcal{N}(0,1), t} \left[ \left\| \epsilon - \epsilon_\theta(z_t, t, e_g) \right\|_2^2 \right],
\end{equation}
where \(x\) is a sampled human image, \(t\) denotes the sampled de-noising step, and \(e_g\) represents the guidance embedding generated by our condition module.
We compute the MSE  loss between the sampled noise $\epsilon$ and the estimated noise $\epsilon_\theta(z_t, t, e_g)$ for optimization.

\begin{figure}[t]
\centering
\includegraphics[width=0.48\textwidth]{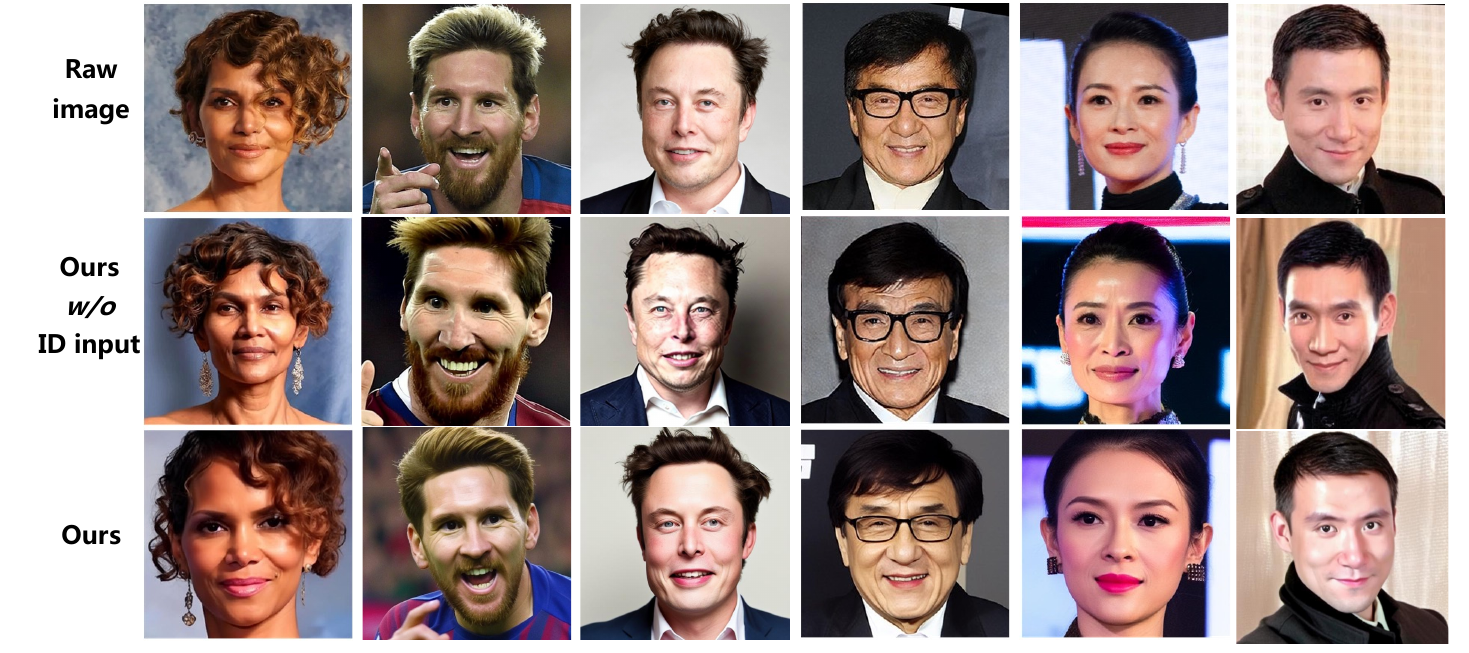}
\vspace{-0.5 em}
\caption{\textbf{Influence of identity input on image construction.}  The addition of identity input proves to be effective in preserving the subject's identity within the generated image.} 
\label{fig:reconstruction}
\vspace{-0.5 em}
\end{figure}

\begin{figure}[t]
\centering
\includegraphics[width=0.45\textwidth]{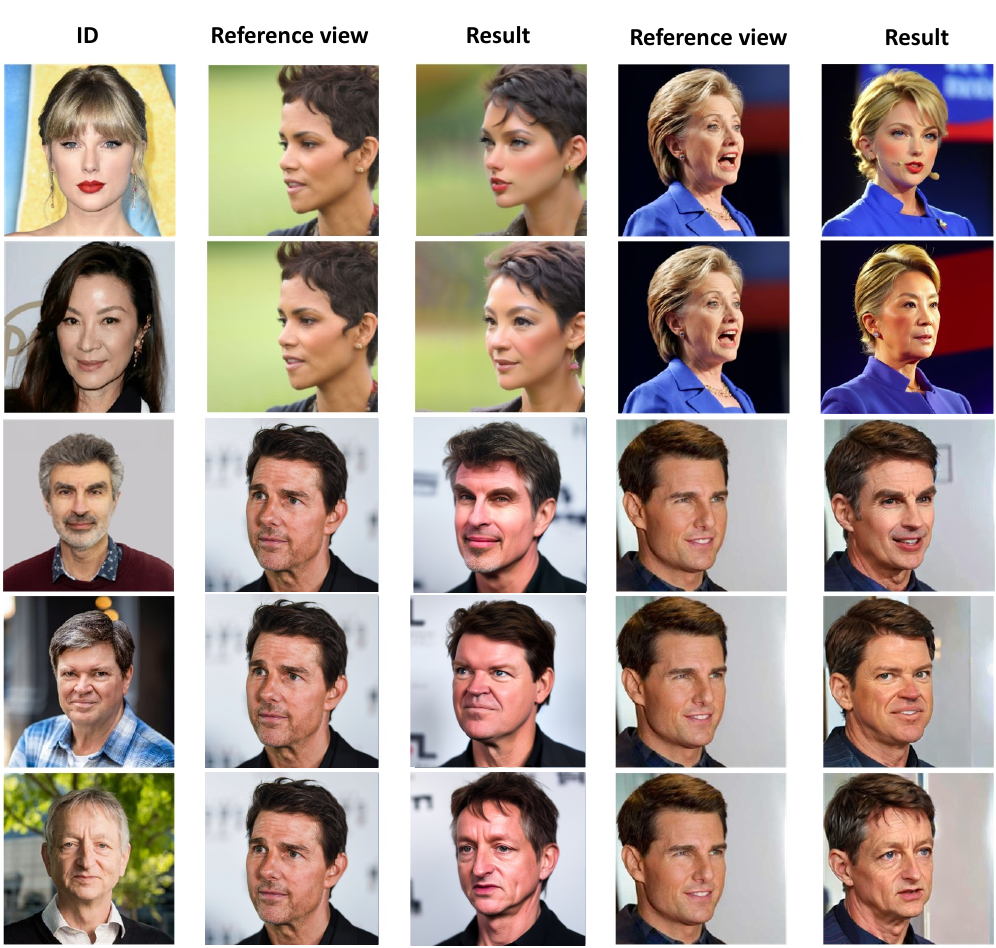}
\vspace{-0.5 em}
\caption{\textbf{Identity-preserving novel view synthesis experiment.} Our method excels at generating new views of a subject while maintaining its identity.  } 
\label{fig:novel_view}
\vspace{-1 em}
\end{figure}

\begin{figure*}[t]
\centering
\includegraphics[width=0.95\textwidth]{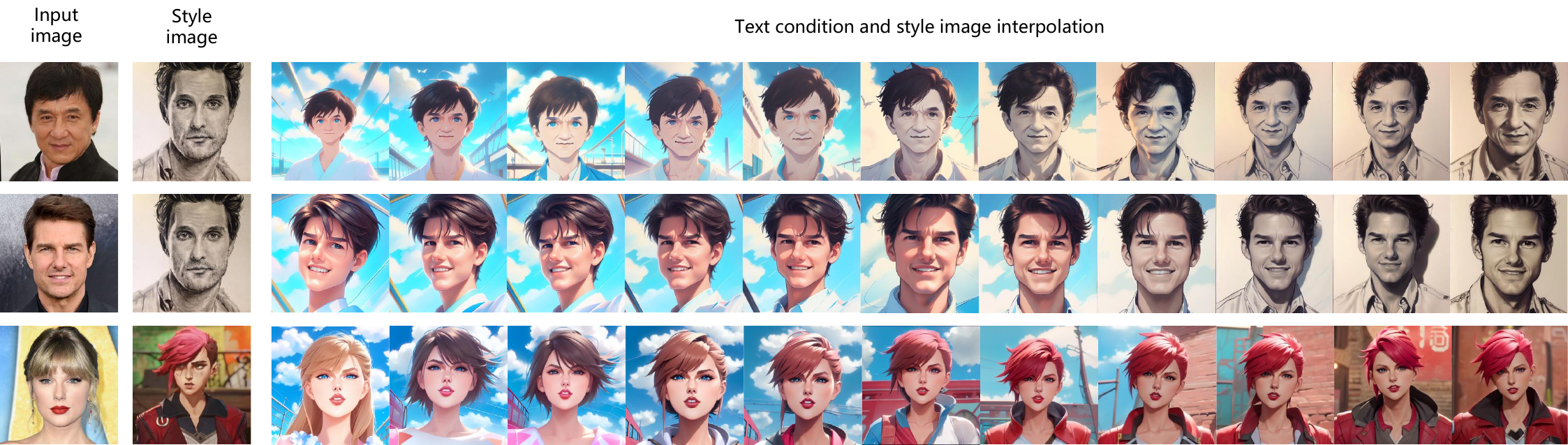}
\vspace{-0.5 em}
\caption{\textbf{Hybrid-guidance experiments.   }  In this experiment, we employ an approach that combines textual prompts and reference images for image synthesis, and the text prompt used here pertains to the cartoon style.  } 
\label{fig:hybrid}
\end{figure*}

\begin{figure*}[t]
\centering
\includegraphics[width=0.98\textwidth]{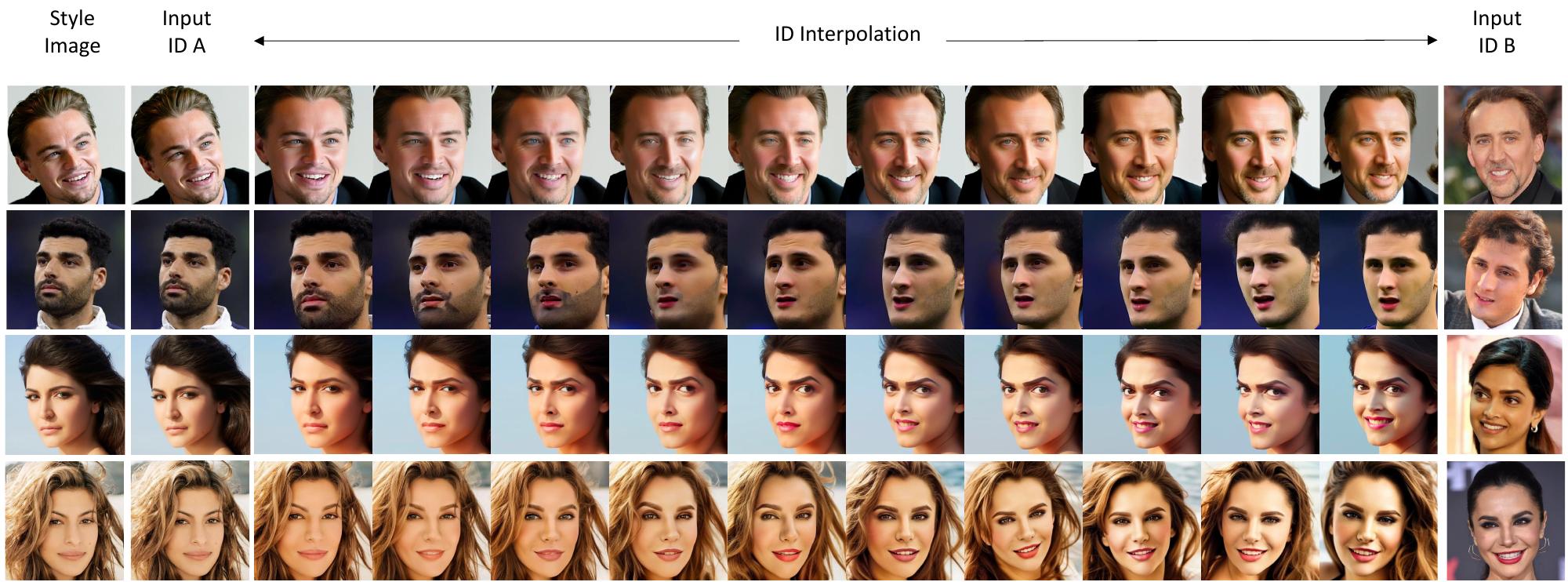}
\vspace{-0.5 em}
\caption{\textbf{Identity mixing experiment.} We generate facial images that combine multiple identities using a mixing ratio to control the influence of different IDs.  } 
\label{fig:mixing}
\vspace{-1 em}
\end{figure*}

\begin{figure*}[t]
\centering
\includegraphics[width=0.95\textwidth]{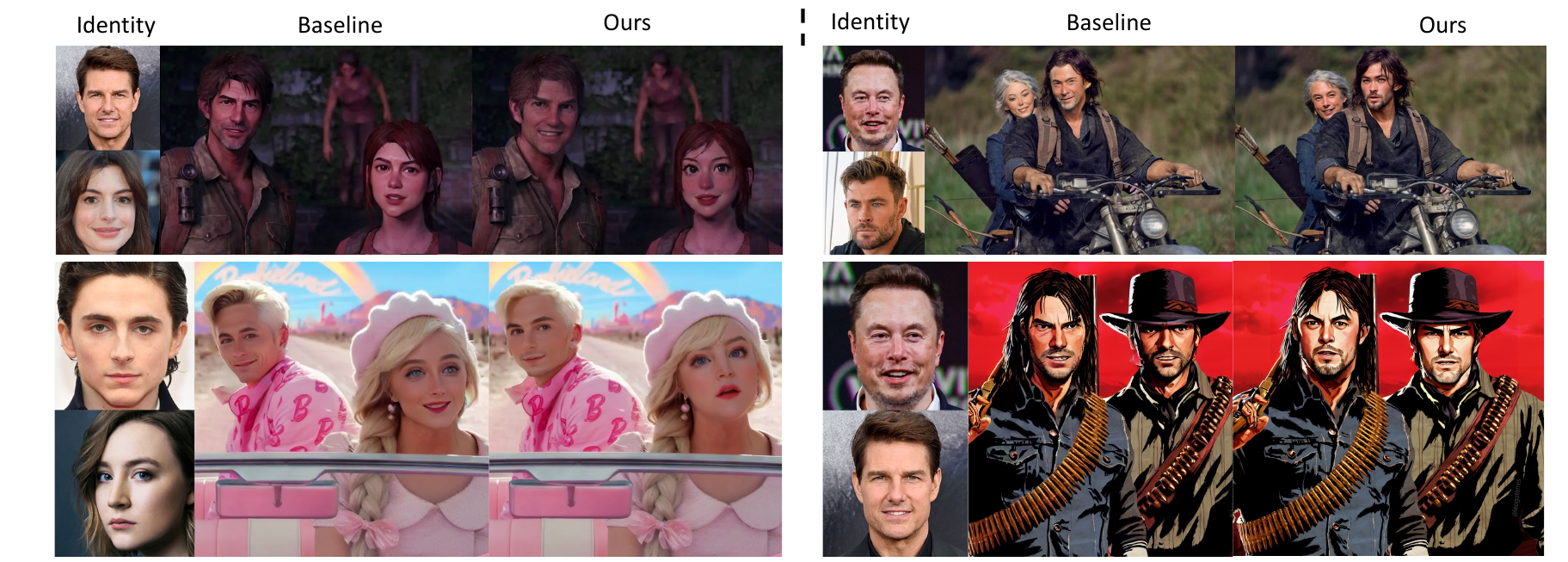}
\vspace{-0.5 em}
\caption{\textbf{Comparison of multi-human image synthesis.}  Our model's effectiveness is evident when compared to our model variant that removes our proposed multi-human cross-attention mechanisms.  } 
\label{fig:multihuman}
\end{figure*}

\begin{figure*}[t]
\centering
\includegraphics[width=0.95\textwidth]{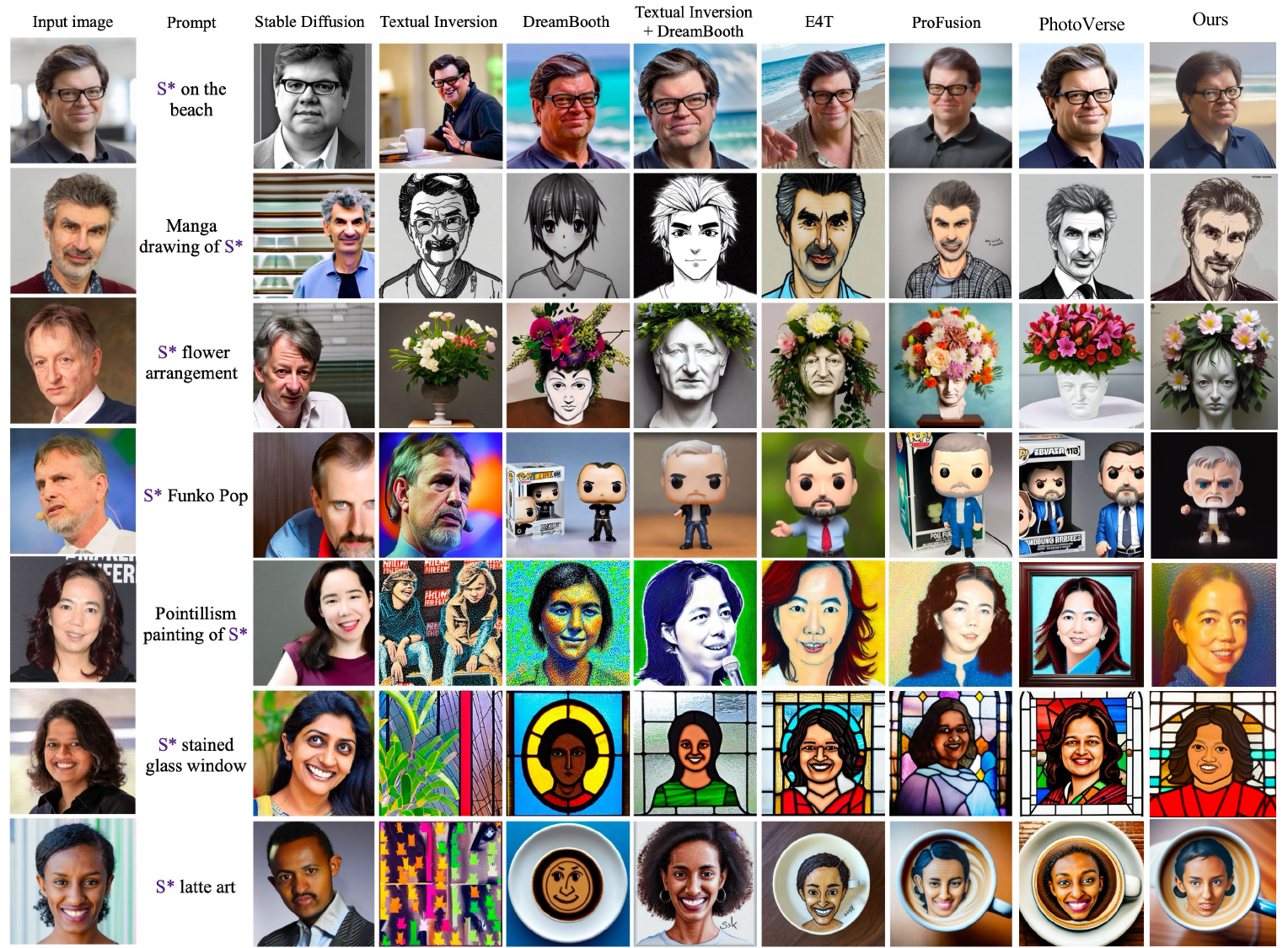}
\caption{\textbf{Comparison with state-of-the-art methods in  identity-preserving text-to-image generation.} This figure illustrates a comparative analysis of our model against state-of-the-art methods in the task of identity-preserving text-to-image generation. The evaluation is conducted using the same samples employed in FaceNet's experiments. Notably, our model consistently achieves comparable or superior qualitative results across all the examples.  } 
\label{fig:sota}
\end{figure*}

\begin{table*}[]
\small
\centering
\begin{tabular}{l|ll|lll}
\toprule[1pt]
\multirow{2}{*}{} & \multicolumn{2}{c|}{\textbf{Face Similarity} $\uparrow$} & \multicolumn{3}{c}{\textbf{Time} (s) $\downarrow$}    \\ \cline{2-6} 
                  & \textbf{Single-Image }     & \textbf{Multi-Image}      & \textbf{Tuning} & \textbf{Inference}    & \textbf{Sum }\\ \hline
DreamBooth~\cite{ruiz2023dreambooth}        & 0.65              & 0.58             & 405    & 2.8         &  407.8    \\
Textual Inversion~\cite{gal2022image} & 0.31              & 0.27             & 3425   & 2.8         &   3427.8   \\ \hline
Ours \textit{w/o} ID input        & \multicolumn{2}{c|}{0.47}            & 0      & 3.6          &  3.6    \\ \toprule[1pt]
\textbf{Ours \textit{w/} text }       & \textbf{0.71}              & \textbf{0.61}             & 0      & \textbf{5.3 + 0.04 N}  &  \textbf{5.3 + 0.04 N}   \\
\textbf{Ours \textit{w/} image }        & \textbf{0.86 }             & \textbf{0.74  }           & 0      & \textbf{3.6 + 0.04 N} &   \textbf{3.6 + 0.04 N}  \\ \toprule[1pt]
\end{tabular}
\vspace{-0.5 em}
\caption{\textbf{Comparison of face similarity and generation time for identity-preserving image generation.} 
Our methods, guided by both texts and images, exhibit remarkable advantages compared to baseline approaches in terms of both face similarity and generation time. The variable \textbf{N} represents the number of reference images per identity. Notably, the omission of identity guidance input in our design results in a substantial drop in performance.
}
\label{tab:similarity}
\vspace{-1.5 em}
\end{table*}

\begin{figure*}[t]
\centering
\includegraphics[width=0.95\textwidth]{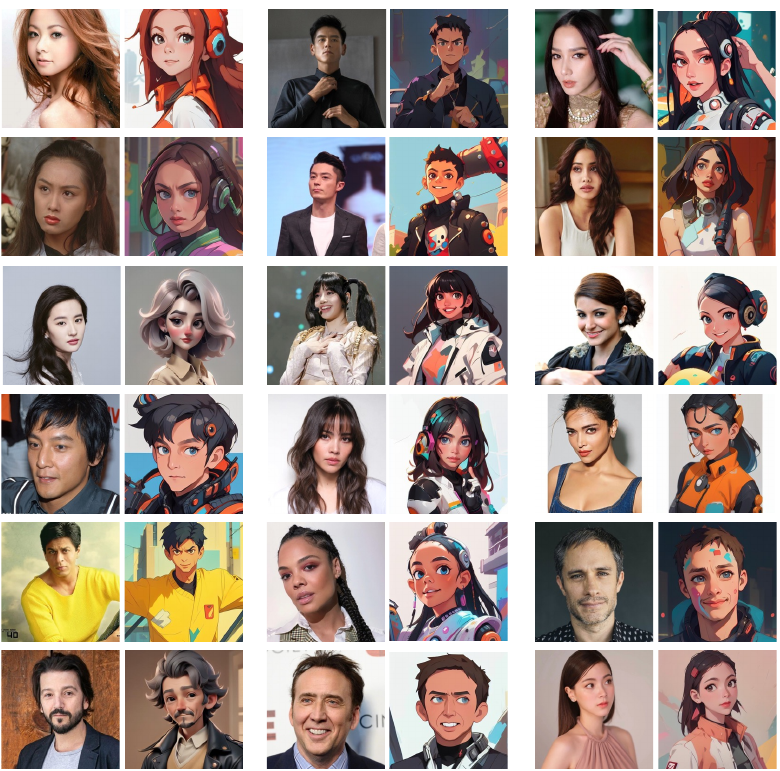}
\caption{\textbf{Image-to-image synthesis with our proposed method.  }  Our model preserves the identities of humans and the layout in the raw images.  } 
\label{fig:i2i}
\end{figure*}

\begin{figure*}[t]
\centering
\includegraphics[width=0.85\textwidth]{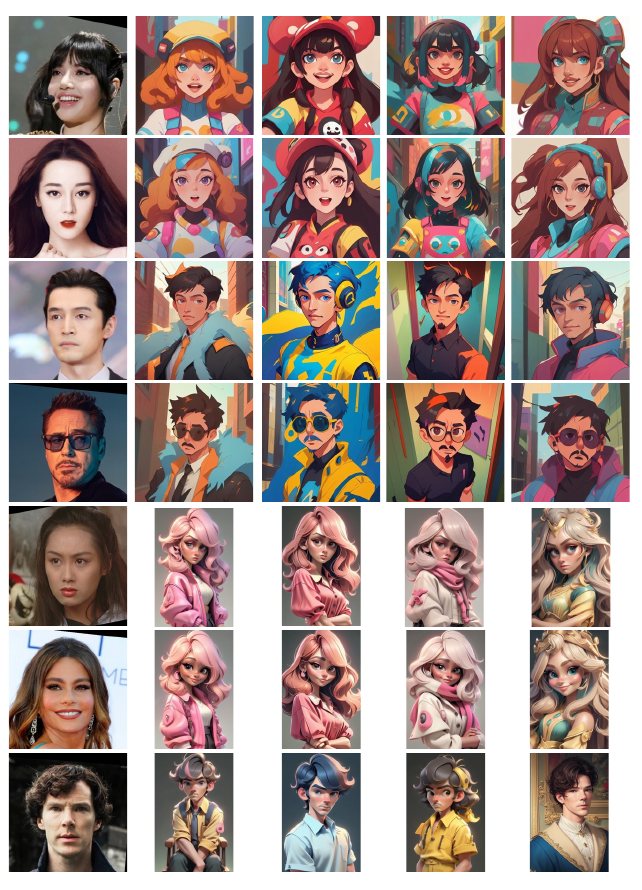}
\caption{\textbf{More qualitative results.   } Our model obtains a balance between stylistic expression and the need to maintain recognizable features of the subject. } 
\label{fig:more}
\end{figure*}

\section{Experiments}

\subsection{Implementation details.}
The vision encoder utilized in the image-conditioned branch of our model combines three CLIP model~\cite{clip} variants with different backbones. These are: CLIP-ViT-L/14, CLIP-RN101, and CLIP-ViT-B/32. The outputs from these individual models are concatenated to produce the final output of our vision encoder.
Our approach primarily utilizes the DDPM configuration~\cite{ddpm} as described in StableDiffusion~\cite{stablediffusion} for training. Specifically, we incorporated a total of 1,000 denoising steps. For the inference stage, we use the EulerA sampler~\cite{eulera} and set it to operate over 25 timesteps.
To align with the training methodology of classifier-free guidance~\cite{ho2022classifier}, we introduced variability by randomly omitting the conditional embeddings related to both style images and face images. Specifically, the probabilities for dropping these embeddings were set at 0.64 for style images and 0.1 for face images.

The primary dataset used for training was FFHQ~\cite{karras2018stylebased}, which is a face image dataset encompassing 70,000 images. To augment this, we also incorporated a subset of the LAION dataset~\cite{schuhmann2022laion5b} into our training phase, which aims to ensure the model retains the capability to generate generic, non-human images during the finetuning process. It's worth noting that when non-human images are sampled for training, the face embedding in the conditional branch is set to zero. During training, we set the learning rate at 1e-6. The model was trained using 8 A100 GPUs, with a batch size of 256, and was trained for 100,000 steps.

\subsection{Results}

We provide quantitative and qualitative results for comparison and analysis in this section.

\textbf{Quantitative comparison with baselines.} 
To evaluate our model's ability to preserve identity during image generation, we conduct a quantitative comparison with baselines. We measure the Arcface feature cosine similarity~\cite{deng2018arcface} between the face regions of input reference human images and the generated images, with values in the range of (-1, 1). We consider both 1-shot and multi-shot cases, where only one reference image and 11 reference images are provided, respectively.
The baselines we compare against include DreamBooth~\cite{ruiz2023dreambooth} and Textual Inversion~\cite{gal2022image}, which are the most popular tuning-based methods, which are optimized with 1K and 3K iterations, respectively. We also incorporate our model variant that removes the identity input for comparison to measure the influence of identity information. 
For the multi-shot experiment, we compute the face similarity score between the generated image and each reference image and report the mean score for comparison.
We have conducted experiments on 11 different identities, and we report the average performance for comparison, and the results are shown in Table~\ref{tab:similarity}.
Our model demonstrates superior performance in both 1-shot and multi-shot cases, highlighting the effectiveness of our design in preserving identities. Notably, our advantages stem from the inclusion of our identity-guidance branch, as demonstrated by the performance drop in the baselines that lack this guidance.

\textbf{  Influence of the identity input.}
To further investigate the influence of the identity input, we conducted an ablation study through qualitative comparisons. We compare our model with a baseline model that removes the identity-guidance branch,  acting as an image reconstruction model.  The results are displayed in Fig.~\ref{fig:reconstruction}.
The image reconstruction baseline roughly preserves image content but struggles with fine-grained identity information. In contrast, our model successfully extracts identity information from the identity-guidance branch, resulting in improved results for the face region.

\textbf{Novel view synthesis.}
We conduct novel view synthesis experiments to validate our algorithm's effectiveness in synthesizing images with large pose changes while preserving identity. Results are presented in Fig.~\ref{fig:novel_view}. The result demonstrates that our algorithm can synthesize high-quality images with preserved identities even when faced with significant pose changes, which showcases the robustness of our design.

\textbf{Dual guidance experiment.}
We explore the combined use of text-based and image-based guidance during inference by controlling the strength of both types of guidance using the hyperparameter, $\alpha$, within the range [0,1]. The results, presented in Fig.~\ref{fig:hybrid}, illustrate that our model can effectively utilize both types of guidance for image synthesis while preserving identity information. By adjusting $\alpha$, we could see how the influence of each type of guidance changed.

\textbf{Identity mixing experiment.} 
Our model's ability to mix identity information from different humans during image synthesis is showcased in Fig.~\ref{fig:mixing}. By controlling the mix ratio within the range [0,1], we assign weights to different identities. As is shown, our model effectively combines identity information from different people and synthesizes new identities with high fidelity.

\textbf{Multi-human image generation. }
One of our model's unique features is synthesizing multi-human images from multiple identities. We present the results in Fig.~\ref{fig:multihuman}, by comparing our design to a baseline using vanilla cross-attention mechanisms. As the result shows, our model effectively correlates different human regions with different identities with our proposed enhanced cross-attention mechanisms to differentiate between identities, while the baseline results in confused identities in the human regions.

\textbf{More qualitative results.}
Fig.\ref{fig:i2i} showcases a new image synthesis method, which is an extension of the image-to-image generation technique originally found in StableDiffusion\cite{stablediffusion}. This method has only minor variations from the primary pipeline discussed in the main paper. In this approach, the diffusion process begins with the noisy version of the raw human images' latent representations, with the rest of the process unchanged. This specific modification ensures that the synthesized images retain a layout similar to the original images.
Our results demonstrate that, despite these adjustments, the method successfully maintains the identity of subjects in the synthesized images. 
Additionally, Fig.~\ref{fig:more} provides more qualitative results of our model when applied to a broader range of image styles. These results highlight the model's adaptability to various artistic styles while still holding true to the core objective of identity preservation. This versatility is crucial for applications that require a balance between stylistic expression and the need to maintain recognizable features of the subject.

\section{Conclusion}

In this paper, we present an innovative approach to text-to-image generation, specifically focusing on preserving identity in the synthesized images. Our method significantly accelerates and enhances the efficiency of the image generation process. 
Central to our approach is the hybrid guidance strategy, which combines stylized and facial images with textual prompts, guiding the image generation process in a cohesive manner. A standout feature of our method is its ability to synthesize multi-human images, thanks to our developed multi-identity cross-attention mechanisms. Our extensive experimental evaluations, both qualitative and quantitative, have shown the advantages of our method. 
It surpasses baseline models and previous works in several key aspects, most notably in efficiency and the ability to maintain identity integrity in the synthesized images.

\textbf{Limitation and Social Impacts.}
Compared to existing works like DreamBooth~\cite{ruiz2023dreambooth}, which synthesize images of diverse subjects such as animals and objects, our model is specifically tailored for identity-preserving generation, exclusively targeting human images.
Our text-to-image generation research has two key societal impacts to consider: 1) Intellectual Property Concerns. The ability of our model to create detailed and stylized images raises potential issues with copyright infringement. 2) Ethical Considerations in Facial Generation. The model's capability to replicate human faces brings up ethical issues, especially the potential for creating offensive or culturally inappropriate images. It's crucial to use this technology responsibly and establish guidelines to prevent its misuse in sensitive contexts.

\clearpage

{
    \small
    \bibliographystyle{ieee_fullname}
    \bibliography{egbib}
}

\end{document}